\documentclass[letterpaper]{article}
\usepackage{spconf,amsmath,graphicx}
\usepackage{adjustbox}
\usepackage{multirow}
\usepackage{amsmath}
\usepackage{amssymb}
\usepackage[table]{xcolor}
\usepackage{mathrsfs}
\usepackage{dsfont}
\usepackage{bm}

\newcommand{\xu}[1]{{\color{black}{#1}}}

\title{VoteNet+ : An Improved Deep learning label fusion method for multi-atlas segmentation}
\name{Zhipeng Ding$^1$, Xu Han$^1$, Marc Niethammer$^{1,2}$}
\address{$^1$Department of Computer Science, UNC Chapel Hill, USA\\
$^2$Biomedical Research Imaging Center, UNC Chapel Hill, USA\\}
\begin{document}
%
\maketitle
\begin{abstract}
In this work, we improve the performance of multi-atlas segmentation (MAS) by integrating the recently proposed VoteNet model with the joint label fusion (JLF) approach. Specifically, we first illustrate that using a deep convolutional neural network to predict atlas probabilities can better distinguish correct atlas labels from incorrect ones than relying on image intensity difference as is typical in JLF. Motivated by this finding, we propose VoteNet+, an improved deep network to locally predict the probability of an atlas label to differ from the label of the target image. Furthermore, we show that JLF is more suitable for the VoteNet framework as a label fusion method than plurality voting. Lastly, we use Platt scaling to calibrate the probabilities of our new model. Results on LPBA40 3D MR brain images show that our proposed method can achieve better performance than VoteNet. 
\end{abstract}
\section{Introduction}
\label{sec:introduction}
Image segmentation, i.e. assigning pixel-wise or voxel-wise labels, is important for image-based diagnosis and analysis~\cite{iglesias2015multi}. Thus much effort has been spent on developing fast and accurate segmentation algorithms. Recently, deep-learning (DL) approaches~\cite{cciccek20163d,oktay2018attention,abraham2019novel} have started to dominate the field of medical image analysis for many tasks, including image segmentation, due to their good performance.

Prior to the ascent of DL approaches, multi-atlas segmentation (MAS) techniques have been widely successful. MAS utilizes deformable image registration to transfer atlas labels to a target image to be segmented. As label maps are deformed in a controlled manner via deformable registration approaches local structure and topology stay well behaved in the target image space, thus helping to retain spatial consistency (e.g., to avoid adding unrealistic structures or missing structures). However, besides these desirable MAS behaviors, MAS is slow as it relies on costly image registrations and has been outperformed for many segmentation tasks by DL approaches. Recently, VoteNet~\cite{ding2019votenet} proposed an approach to use DL to predict trust-worthy atlases for MAS and to use DL-based image registration. This allowed matching DL performance while retaining spatial consistency. 

\noindent
\textbf{Related Work.} There are various recent approaches combining MAS with machine learning. E.g., \cite{wang2014multi} proposed to use random forests for patch-based label fusion. Further, \cite{ding2019votenet} uses a CNN to locally select the best atlases and \cite{xie2019improving} uses a CNN to predict patch-based similarity in the JLF framework. Most related to our work are~\cite{ding2019votenet,xie2019improving}.
However, our approach differs in the following ways: a) we design VoteNet+, a network which improves over VoteNet in~\cite{ding2019votenet}; b) we use JLF  for label fusion while~\cite{ding2019votenet} use plurality voting; c) we predict the probability for an entire image instead of focusing on patch centers as in~\cite{xie2019improving}; d) we propose to use Platt scaling to correct probabilities predicted by our CNN while~\cite{xie2019improving} use a heuristic to approximate probabilities. Sec.~\ref{sec:experiment_result_discussion} demonstrates that our proposed approach can indeed achieve improved performance.

\noindent
\textbf{Contributions.} (1)~\emph{New network architecture}: We propose a new deep convolutional network (VoteNet+), which locally identifies sets of trustworthy atlases more accurately than VoteNet. 
(2)~\emph{Probability Calibration}: We calibrate the probabilities of our network resulting in more accurate segmentations. (3) We show the advantages of combining VoteNet and Joint Label Fusion via Oracle experiments. (4) We further improve the final segmentation performance by combining VoteNet+ with a U-Net-based segmentation network.
   

\begin{figure*}[t]
\centering
  \includegraphics[width=0.9\linewidth]{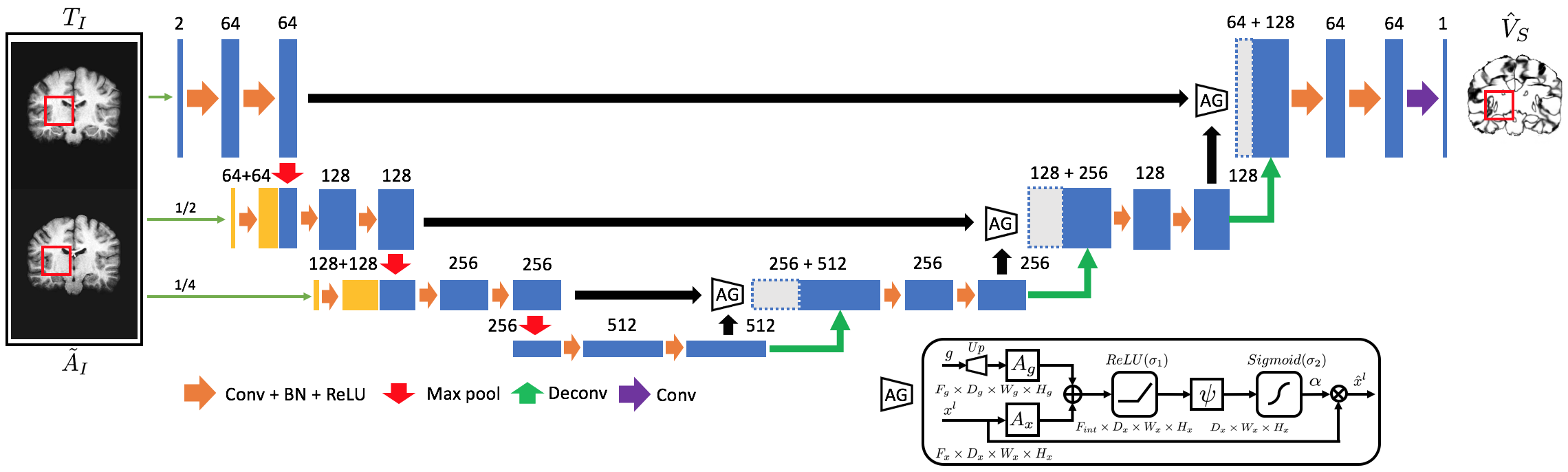}
  \caption{VoteNet+ architecture. We use a customized attention U-Net with multi-scale input. Feature dimensions (channels) of each convolutional layer are listed on the top of each block. VoteNet+ takes the target image and a warped atlas image as inputs and outputs a voxel-wise probability indicating how likely the label from the warped atlas image is correct for the target image.}
  \label{fig:framework}
  \vspace{-0.5ex}
\end{figure*}

\section{Methodology}
\label{sec:method}
\subsection{MAS Overview}
\label{sec:mas_overview}
Assume we have $n$ atlas images and their corresponding manual segmentations: $A^1 = (A^1_I, A^1_S), A^2 = (A^2_I, A^2_S), ..., A^n = (A^n_I, A^n_S)$. Let $T_I$ be the target image that needs to be segmented. MAS first uses a reliable deformable image registration method to warp all atlases to the target image space, i.e. $\tilde{A}^1 = (\tilde{A}^1_I, \tilde{A}^1_S), \tilde{A}^2 = (\tilde{A}^2_I, \tilde{A}^2_S), ..., \tilde{A}^n = (\tilde{A}^n_I, \tilde{A}^n_S)$; and then uses a label fusion method, $\mathscr{F}$, to combine all the candidate segmentations to produce the final segmentation $\hat{T}_S$ for $T_I$, i.e.
\begin{equation}
\label{eq:est}
\hat{T}_S = \mathscr{F}(\tilde{A}^1, \tilde{A}^2, ..., \tilde{A}^n, T_I).
\end{equation}

\subsection{Joint Label Fusion}
\label{sec:joint_label_fusion}
Joint Label Fusion (JLF)~\cite{wang2012multi} is a label fusion method taking into account correlated errors of atlases. It models binary segmentations, but can be extended to multi-label settings. The following is a brief introduction.

Binary segmentation errors can be modeled as
\begin{equation}
T_S(\bm{x}) = \tilde{A}^i_S(\bm{x}) + \delta^i (\bm{x})
\end{equation}
where $T_S(\bm{x})$ and $\tilde{A}^i_S(\bm{x})$ are the true target segmentation and the $i$th atlas segmentation at position $\bm{x}$; $\delta^i(\bm{x})$ is the label difference. When $T_S(\bm{x}) = 1$, $\delta^i(\bm{x}) \in \{0, 1\}$; otherwise when $T_S(\bm{x}) = 0$, $\delta^i(\bm{x}) \in \{-1, 0\}$. The consensus segmentation $\hat{T}_S$ is modeled as $\hat{T}_S (\bm{x}) = \sum_{i=1}^n \omega_i (\bm{x}) \tilde{A}^i_S (\bm{x})$, where $0\leq\omega_i (\bm{x})\leq 1$ is the weight assigned to the $i$th atlas and $\sum_{i=1}^n \omega_i (\bm{x}) = 1$. JLF tries to find the optimal weights $\omega_i (\bm{x})$ minimizing the expected error between $\hat{T}_S (\bm{x})$ and the true segmentation $T_S (\bm{x})$:
\begin{equation}
E \left [ (T_S (\bm{x}) - \hat{T}_S (\bm{x}))^2  \right ].
\end{equation}
Ignoring that $\omega_i(x)\in[0,1]$ the JLF weights can be computed in closed-form as
\begin{equation}
\textbf{w}_x = \frac{\textbf{M}^{-1}_{x} \textbf{1}_n }{\textbf{1}^t_n \textbf{M}^{-1}_{x} \textbf{1}_n}
\end{equation}
where $\textbf{w}_x$ is the weight vector that contains weight of all atlases. $\textbf{M}_x\in\mathbb{R}^{n\times n}$ is the dependency matrix, whose entries are the joint probabilities of both the $i$th atlas (row) and the $j$th atlas (column) producing the wrong label for the target image, i.e.,
\begin{equation}
\begin{split}
\textbf{M}_x (i, j) = p(\delta^i (\bm{x})\delta^j(\bm{x}) = 1).
\end{split}
\end{equation}

\subsection{Network Architecture}
\label{sec:network_architecture}
Designing a good network to predict the probability of whether an atlas locally has the same label as the target image is a key aspect of the VoteNet~\cite{ding2019votenet} framework. Still building on top of the 3D-Unet~\cite{cciccek20163d}, we improved the network architecture (VoteNet+) to better predict such probabilities. We use an image pyramid to provide structure details at different scales. This image pyramid is injected into the second and third encoder blocks and concatenated with maxpooling features from previous encoding blocks, as illustrated in Fig.~\ref{fig:framework}(left). Further, inspired by~\cite{oktay2018attention,abraham2019novel}, we use customized soft attention gates (AGs) to help identify where label mis-assignment might occur. AGs produce attention coefficients $\alpha \in [0, 1]$ at each voxel to scale the input feature maps $x^l$ of layer $l$ to output salient features $\hat{x}^l$. As illustrated in Fig.~\ref{fig:framework}(bottom right), a gating tensor $g$, which is used to determine focus regions, is first upsampled to the same shape as the features $x^l$. Additive attention is then formulated as follows:
\begin{equation}
\begin{split}
\centering
q_{att}^l &= \psi^T \left ( \sigma_1 (A_x^T x^l + A_g^T g + b_g ) \right ) + b_{\psi} \\
\alpha^{l} &= \sigma_2 (q_{att}^l(x^l, g; \Theta_{att})) \\
\hat{x}^l &= \alpha^l \odot x^l 
\end{split}
\end{equation}
where $\sigma_2$ is a sigmoid function; $\sigma_1$ is a ReLU activation function; linear transformations $A_x$, $A_g$ and $\psi$ are computed using channel-wise $1\times1\times1$ convolutions; $b_g$ and $b_\psi$ are bias terms; all the parameters in AG are represented as $\Theta_{att}$.

Let the network be $\textbf{P}$. Inputs are the target image $T_I$ and the warped atlas image $\tilde{A}_I$; the output of the network is the probability $p(\tilde{A}_I = T_S)$. Thus, to approximate the joint probability in Sec.~\ref{sec:joint_label_fusion}, we have 
\begin{equation}
\label{eq:probability}
\begin{split}
p(\delta^i \neq 0) &= 1 - p(\tilde{A}^i_I = T_S) = 1 - \textbf{P}(T_I, \tilde{A}^i_I), \\
p(\delta^i\delta^j = 1) & \approx p(\delta^i \neq 0)p(\delta^j \neq 0),  \\
& = (1 - p(\tilde{A}^i_I = T_S))(1 - p(\tilde{A}^j_I = T_S)), \\
& = (1 - \textbf{P}(T_I, \tilde{A}^i_I))(1 - \textbf{P}(T_I, \tilde{A}^j_I)).
\end{split}
\end{equation}


\subsection{Probability Calibration}
\label{sec:probability_calibration}
Probabilities predicted from deep convolutional neural networks are often not well-calibrated~\cite{guo2017calibration}. We use Platt scaling~\cite{platt1999probabilistic,niculescu2005predicting} to calibrate the probabilities of our deep neural network. Specifically, we fixed the learned parameters of VoteNet+ (Sec.~\ref{sec:network_architecture}) and optimized two scalar parameters $a, b \in \mathds{R}$ and output $\hat{p} = \sigma (az + b)$ as the calibrated probability. Here $\sigma$ is the sigmoid function, $z$ is the output logit of VoteNet+ before going into the sigmoid function. Parameters $a$ and $b$ are optimized using the negative log likelihood (NLL) loss over the validation dataset of VoteNet+.

\section{Experiment \& Implementation details}
\label{sec:implementation}
\textbf{Dataset.} We use 40 3D MR brain images from the LONI Probabilistic Brain Atlas (LPBA40) dataset. Each image contains 56 manually segmented structures excluding background. All images are affine registered to the MNI152 atlas and histogram equalized. The dataset is randomly divided into two non-overlapping equal-size subsets for two-fold cross-validation. 17 images of the training dataset are chosen as atlases and the remaining 3 for validation. All results presented in Tab.~\ref{tab:metrics} are averaged over the two folds.

\noindent
\textbf{VoteNet+.} The network takes patches of size $72\times72\times72$ from the target image and a warped atlas image at the same position, where the $40\times40\times40$ patch center is used to tile the volume. In the training dataset, all 17 atlases are registered to the other 19 images using Quicksilver~\cite{yang2017quicksilver}, which results in $17\times19$ pairs. Output is the voxel-wise probability that indicates whether the warped atlas label is equal to the target image label. Binary cross entropy loss is used as the loss function. We train using \texttt{ADAM} over 500 epochs with a multi-step learning rate. The initial learning rate is 0.001 and is reduced by half at the 200th epoch, 350th epoch, and 450th epoch respectively. Training patches are randomly cropped assuring at least 5\% correct labels in the patch volume. Training takes $\approx$58 hours on an NVIDIA GTX1080Ti and testing for a single atlas image takes less than 20 seconds.   

\noindent
\textbf{Platt Scaling.} We train using \texttt{ADAM} over 2,000 epochs with a fixed learning rate of 0.00001. NLL is only calculated inside the brain area via a binary mask created using the validation dataset to remove the influence of background voxels. Training takes $\approx$5 mins on an NVIDIA GTX 1080Ti and the learned parameters are $a = 0.733$ and $b = 0.049$.

\section{Results and Discussion}
\label{sec:experiment_result_discussion}

\noindent
\textbf{Metrics.} We use Dice scores to evaluate the performance of different methods. Due to the length constraints of the manuscript, we report Dice scores on the most difficult region (right lateral orbitofrontal gyrus) and the easiest region (cerebellum). We also include average Dice scores over all 56 regions as an overall performance measure.

\begin{figure}[h]
\centering
  \includegraphics[width=0.9\linewidth]{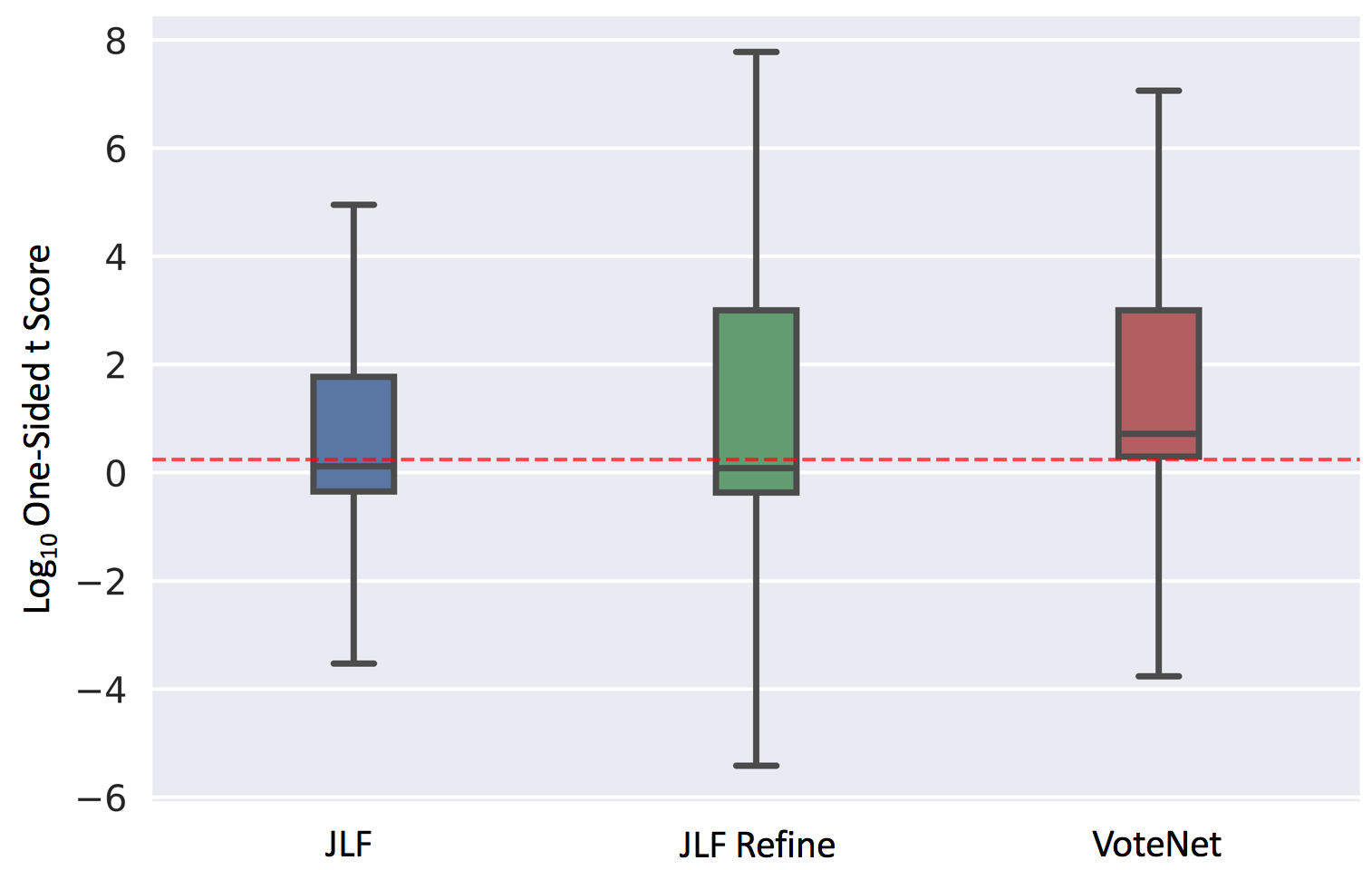}
  \caption{t-scores for three methods. \textbf{JLF}: Joint Label Fusion without local patch search refinement; \textbf{JLF Refine}: Joint Label Fusion with local patch search refinement; \textbf{VoteNet}: VoteNet from~\cite{ding2019votenet}. The red line approximately indicates the 5\% significance level. VoteNet using CNN predicted probabilities significantly outperforms JLF using image intensities (with or without refinement).}
  \label{fig:t_score}
  \vspace{-0.5ex}
\end{figure}

\noindent
\textbf{One-sided t score.} In~\cite{wang2012multi}, the probability of the $i$th atlas having the wrong label at position $x$ is defined as
\begin{equation}
p(\delta^i(\bm{x}) \neq 0) = \left [ \sum_{\bm{y} \in \mathcal{N}(\bm{x})} (T_I(\bm{y}) - \tilde{A}^i_I (\bm{y}) )^2\right ]^{\beta}\label{eq:p_jlf}
\end{equation}
while in VoteNet, it is simply
\begin{equation} 
p(\delta^i(\bm{x}) \neq 0) = 1 - \textbf{P}(T_I(\bm{x}), \tilde{A}^i_I(\bm{x})).\label{eq:p_votenet}
\end{equation}
To compare which approach is better at distinguishing correct labels from incorrect ones we compute a one-sided t-score. Specifically, given one target image and its corresponding manual segmentation, at each voxel, we record which atlases are correct or not with their associated probabilities from equations~\ref{eq:p_jlf}\footnote{In the experiment, $\beta$=1, $\mathcal{N}(\bm{x})$ is a $5\times5\times5$ cube.} and~\ref{eq:p_votenet}. Hence, for each voxel, we obtain a set of correct and a set of incorrect atlases. We then locally compute the one-sided t-score for the probability values of the correct versus the incorrect set. A positive t-score\footnote{Some values will be infinite if sets are empty or only contain one element. For example, if all 17 atlases assign the incorrect label, the value would be $+\infty$. To simplify the visualization, we clamp the t-scores to -1000($-\infty$ or all wrong) or 1000($+\infty$ or all correct).} indicates that the incorrect set has a mean higher than the correct set, which is expected. Note that the absolute value of the t-score measures how far away the distributions of the two groups are from each other, so higher is better under this measure. Note that in~\cite{wang2012multi}, there is an additional local patch search step to mitigate image registration errors and to refine the JLF results. Thus, we compare the t-scores of three methods: JLF without refinement, JLF with refinement, and VoteNet. Fig.~\ref{fig:t_score} shows the results. We observe that JLF with refinement enlarges the range of differences between the distributions of the correct and incorrect assignments over JLF without refinement. VoteNet using a CNN to predict the probabilities better distinguishes the distributions (i.e., results in overall higher t-scores).

\noindent
\textbf{Oracle results.} Given probabilities (from equation~\ref{eq:p_votenet}), we show that JLF is better than plurality voting (used in~\cite{ding2019votenet}) as a label fusion method. Specifically, we create 4 Oracle experiments, which assume we know the true segmentation of all images. In the first part of Tab.~\ref{tab:metrics}, we assign 0.4 to equation~\ref{eq:p_votenet} if the voxel-wise warped atlas segmentation is the same as the target image segmentation;
otherwise, we assign 0.6. \textit{G} means adding an independent Gaussian noise with mean 0 and standard deviation 0.2 to the probabilities for each atlas. \textit{GS} means that the same Gaussian noise with mean 0 and standard deviation 0.2 is added to each atlas.
\textit{(P)} means using plurality voting (with a threshold of 0.5) as the label fusion method while \textit{(J)} means using JLF. 
Results on all 4 Oracle experiments show that JLF outperforms plurality voting as the label fusion method. 
Especially in the case when adding the same Gaussian noise to probabilities of each atlas, JLF greatly outperforms plurality voting. This is because JLF only needs to preserve the property that the correct atlas label has lower probability than the incorrect atlas label in equation~\ref{eq:p_votenet}. This is a much weaker requirement than using plurality voting with the assumption that the correct atlas label has low probability and the incorrect atlas label has high probability in equation~\ref{eq:p_votenet}.
Consider the case where all 17 atlases give relatively high probabilities in equation~\ref{eq:p_votenet} (e.g. 0.7 for correct atlas label, 0.9 for incorrect atlas label). In this case plurality voting will not assign a label while JLF results in the correct assignment. Situations are similar for low probabilities. Thus, JLF is more suitable than plurality voting in the VoteNet framework.

The above two experiments consequently motivate us to integrate JLF into the VoteNet(+) framework. 


\textbf{Analysis.} The second and third parts of Tab.~\ref{tab:metrics} contain several comparisons. 1) \textit{Plurality voting vs JLF}: JLF outperforms plurality voting as a stand-alone label fusion method as well as in the VoteNet \xu{and VoteNet+} framework. 2) \textit{VoteNet vs VoteNet+}: For all kinds of combinations, VoteNet+ performs better than VoteNet \xu{both with plurality voting and with JLF}. In fact, VoteNet+ achieves a 2\% improvement for predicting whether an atlas label is equal to the target label or not, \xu{although} in the final label fusion stage, the 2\% improvement only translates to $\approx$0.4\% for the fused segmentation. 
3) \textit{VoteNet\xu{+} vs U-Net}: It was shown in~\cite{ding2019votenet} that using U-Net segmentation results to label voxels for which VoteNet could not find any trustworthy atlases can further improve results. 
Here, we examine the combination of VoteNet\xu{+} with JLF and U-Net. We follow the approach in~\cite{ding2019votenet}, but instead of using plurality voting we use JLF within the VoteNet\xu{+} framework. 
For both VoteNet and VoteNet+, the final segmentations are improved. 4) \textit{Probability Calibration}: We examined applying Platt scaling to correct the probabilities predicted from VoteNet+. We found that after probability calibration, thousands of voxel assignments become correct. However, this resulted in only modest improvements in Dice score.

\begin{table}[th] 
\centering
\normalsize
\begin{adjustbox}{max width=0.45\textwidth}
\begin{tabular}{ cccc}
\hline  
Method  & R LOG  & cerebellum &Average \\ 
\hline 
Oracle G (P) & 80.01 $\pm$ 6.68 & 96.92 $\pm$ 0.81 &86.88 $\pm$ 1.04\\
Oracle GS (P) & 68.38 $\pm$ 3.55 & 79.68 $\pm$ 0.54  &72.93 $\pm$ 0.53\\
Oracle G (J) & 85.62 $\pm$ 5.00 & 97.78 $\pm$ 0.53 & 90.26 $\pm$ 0.81 \\
Oracle GS (J) & 96.07 $\pm$ 2.59 & 99.32 $\pm$ 0.22 & 97.18 $\pm$ 0.45 \\
\hline
\hline
PV & \cellcolor{green!30}65.79 $\pm$ 8.50 & \cellcolor{green!30}94.57 $\pm$ 1.12  &\cellcolor{green!30}77.47 $\pm$ 1.18\\
JLF & 66.95 $\pm$ 8.76 & \cellcolor{green!30}95.47 $\pm$ 0.70 &\cellcolor{green!30}79.50 $\pm$ 1.12\\
U-Net & 68.02 $\pm$ 7.44 & \cellcolor{green!30}96.20 $\pm$ 0.53 &\cellcolor{green!30}80.46 $\pm$ 1.29\\
VoteNet (P) & 68.33 $\pm$ 8.81 & \cellcolor{green!30}96.11 $\pm$ 0.64 & \cellcolor{green!30}80.55 $\pm$ 1.13 \\
VoteNet (P) \& U-Net  & 68.64 $\pm$ 8.72 & \cellcolor{green!30}96.18 $\pm$ 0.51 &\cellcolor{green!30}80.75 $\pm$ 1.11\\
VoteNet (J) & 69.02 $\pm$ 8.62 & \cellcolor{green!30}96.26 $\pm$ 0.43 & 80.88 $\pm$ 1.08\\
VoteNet (J) \& U-Net & 69.05 $\pm$ 8.59 & \cellcolor{green!30}96.35 $\pm$ 0.46 & 81.03 $\pm$ 1.08\\
\hline
\hline
VoteNet+ (P) & 68.92 $\pm$ 8.97 & 96.38 $\pm$ 0.52 & 80.96 $\pm$ 1.12\\
VoteNet+ (P) \& U-Net & 69.14 $\pm$ 8.87 & 96.41 $\pm$ 0.52 & 81.09 $\pm$ 1.09\\
VoteNet+ (J) & 69.61 $\pm$ 8.58 & 96.50 $\pm$ 0.47 & 81.19 $\pm$ 1.08\\
VoteNet+ (J) \& U-Net & 69.51 $\pm$ 8.55 & \textbf{96.57 $\pm$ 0.49} & 81.30 $\pm$ 1.07\\
VoteNet+ (J)-C & \textbf{69.63 $\pm$ 8.60} & 96.49 $\pm$ 0.47 & 81.21 $\pm$ 1.08\\
VoteNet+ (J)-C \& U-Net & 69.55 $\pm$ 8.56 & 96.56 $\pm$ 0.49 & \textbf{81.31 $\pm$ 1.07}\\
\hline
\end{tabular} 
\end{adjustbox}
\caption{Evaluation on LPBA40 dataset. R LOG stands for right lateral orbitofrontal gyrus. (P) is using plurality voting, (J) is using JLF, and C is using probability calibration. 
We use a Mann-Whitney U-test to check for significant differences to VoteNet+ (J)-C \& U-Net. 
We use a significance level of 0.05 and the Benjamini/Hochberg correction~\cite{benjamini1995controlling} for multiple comparisons with a false discovery rate of 0.05. Results are highlighted in green if VoteNet+ (J)-C \& U-Net performs significantly better than the corresponding method.} 
\label{tab:metrics}
\vspace{-0.5ex}
\end{table}

\section{Conclusion and Future Work}
\label{sec:conclusion}
In this work, we explored the integration of VoteNet and Joint Label Fusion. We found that JLF is more suitable in the framework than plurality voting and yields better segmentation performance. Potential future work includes: 1) Since we show that na\"ively combining U-Net and VoteNet can produce even better results, more sophisticated combinations may  further improve the results. 
2) Platt scaling only uses global parameters $a$ and $b$ to calibrate the probability of all voxels simultaneously. The improvement is modest. It would be worth investigating if a more local Platt scaling would allow correcting wrong voxel labels while retaining most of the correct label assignments. 3) VoteNet+ improves binary segmentation results by 2\%, but only by 0.4\% in the final fused segmentation 
Thus, exploring a task-specific network that focuses on challenging voxels (e.g., where only a few atlases give the correct predictions) would be interesting.\\

\noindent
{\bf Acknowledgements} Research reported in this work was supported by the National Institutes of Health (NIH) and  the  National  Science  Foundation  (NSF)  under  award numbers  NSF EECS-1711776 and  NIH 1R41MH118845. The content is solely the responsibility of the authors and does not necessarily represent the official views of the NIH or the NSF.

\pagebreak

\bibliographystyle{IEEEbib}
\bibliography{refs}

\begin{thebibliography}{10}

\bibitem{iglesias2015multi}
Juan~Eugenio Iglesias and Mert~R Sabuncu,
\newblock ``Multi-atlas segmentation of biomedical images: a survey,''
\newblock {\em Medical image analysis}, vol. 24, no. 1, pp. 205--219, 2015.

\bibitem{cciccek20163d}
{\"O}zg{\"u}n {\c{C}}i{\c{c}}ek, Ahmed Abdulkadir, Soeren~S Lienkamp, Thomas
  Brox, and Olaf Ronneberger,
\newblock ``{3D U-Net}: learning dense volumetric segmentation from sparse
  annotation,''
\newblock in {\em International conference on medical image computing and
  computer-assisted intervention}. Springer, 2016, pp. 424--432.

\bibitem{oktay2018attention}
Ozan Oktay, Jo~Schlemper, Loic~Le Folgoc, Matthew Lee, Mattias Heinrich,
  Kazunari Misawa, Kensaku Mori, Steven McDonagh, Nils~Y Hammerla, Bernhard
  Kainz, et~al.,
\newblock ``Attention {U-Net}: Learning where to look for the pancreas,''
\newblock {\em arXiv preprint arXiv:1804.03999}, 2018.

\bibitem{abraham2019novel}
Nabila Abraham and Naimul~Mefraz Khan,
\newblock ``A novel focal {Tversky} loss function with improved attention
  {U-Net} for lesion segmentation,''
\newblock in {\em 2019 IEEE 16th International Symposium on Biomedical Imaging
  (ISBI 2019)}. IEEE, 2019, pp. 683--687.

\bibitem{ding2019votenet}
Z~Ding, X~Han, and M~Niethammer,
\newblock ``{VoteNet:} a deep learning label fusion method for multi-atlas
  segmentation,''
\newblock in {\em Medical Image Computing and Computer Assisted Intervention
  (MICCAI)}, 2019.

\bibitem{wang2014multi}
Hongzhi Wang, Yu~Cao, and Tanveer Syeda-Mahmood,
\newblock ``Multi-atlas segmentation with learning-based label fusion,''
\newblock in {\em International Workshop on Machine Learning in Medical
  Imaging}. Springer, 2014, pp. 256--263.

\bibitem{xie2019improving}
Long Xie, Jiancong Wang, Mengjin Dong, David~A Wolk, and Paul~A Yushkevich,
\newblock ``Improving multi-atlas segmentation by convolutional neural network
  based patch error estimation,''
\newblock in {\em International Conference on Medical Image Computing and
  Computer-Assisted Intervention}. Springer, 2019, pp. 347--355.

\bibitem{wang2012multi}
Hongzhi Wang, Jung~W Suh, Sandhitsu~R Das, John~B Pluta, Caryne Craige, and
  Paul~A Yushkevich,
\newblock ``Multi-atlas segmentation with joint label fusion,''
\newblock {\em IEEE transactions on pattern analysis and machine intelligence},
  vol. 35, no. 3, pp. 611--623, 2012.

\bibitem{guo2017calibration}
Chuan Guo, Geoff Pleiss, Yu~Sun, and Kilian~Q Weinberger,
\newblock ``On calibration of modern neural networks,''
\newblock in {\em Proceedings of the 34th International Conference on Machine
  Learning-Volume 70}. JMLR. org, 2017, pp. 1321--1330.

\bibitem{platt1999probabilistic}
John Platt et~al.,
\newblock ``Probabilistic outputs for support vector machines and comparisons
  to regularized likelihood methods,''
\newblock {\em Advances in large margin classifiers}, vol. 10, no. 3, pp.
  61--74, 1999.

\bibitem{niculescu2005predicting}
Alexandru Niculescu-Mizil and Rich Caruana,
\newblock ``Predicting good probabilities with supervised learning,''
\newblock in {\em Proceedings of the 22nd international conference on Machine
  learning}. ACM, 2005, pp. 625--632.

\bibitem{yang2017quicksilver}
Xiao Yang, Roland Kwitt, Martin Styner, and Marc Niethammer,
\newblock ``Quicksilver: Fast predictive image registration--a deep learning
  approach,''
\newblock {\em NeuroImage}, vol. 158, pp. 378--396, 2017.

\bibitem{benjamini1995controlling}
Yoav Benjamini and Yosef Hochberg,
\newblock ``Controlling the false discovery rate: a practical and powerful
  approach to multiple testing,''
\newblock {\em Journal of the Royal statistical society: series B
  (Methodological)}, vol. 57, no. 1, pp. 289--300, 1995.

\end{thebibliography}

\end{document}